\pdfoutput=1

\documentclass[11pt]{article}

\usepackage[final]{acl}

\usepackage{times}
\usepackage{latexsym}

\usepackage[T1]{fontenc}

\usepackage[utf8]{inputenc}

\usepackage{microtype}

\usepackage{inconsolata}

\usepackage{graphicx}

\usepackage{booktabs}
\usepackage{multirow}
\usepackage{graphicx}  
\usepackage{amsmath}
\usepackage{amssymb}

\title{\textsc{GRADE}: \textbf{G}enerating multi-hop QA and fine-g\textbf{RA}ined \textbf{D}ifficulty matrix for RAG \textbf{E}valuation}

\author{
  Jeongsoo Lee\textsuperscript{1}* \quad
  Daeyong Kwon\textsuperscript{1,2}* \quad
  Kyohoon Jin\textsuperscript{1}\thanks{\ \ Corresponding author.}
  \\
  \textsuperscript{1}DATUMO \quad
  \textsuperscript{2}Graduate School of Culture Technology, KAIST
  \\
  \texttt{\{jeongsoo.lee, kyohoon.jin\}@selectstar.ai}, \texttt{ejmj63@kaist.ac.kr}
  \\
  * Equal contribution
}

\begin{document}
\maketitle
\begin{abstract}

Retrieval-Augmented Generation (RAG) systems are widely adopted in knowledge-intensive NLP tasks, but current evaluations often overlook the structural complexity and multi-step reasoning required in real-world scenarios. These benchmarks overlook key factors such as the interaction between retrieval difficulty and reasoning depth. To address this gap, we propose \textsc{GRADE}, a novel evaluation framework that models task difficulty along two orthogonal dimensions: (1) reasoning depth, defined by the number of inference steps (hops), and (2) semantic distance between the query and its supporting evidence. We construct a synthetic multi-hop QA dataset from factual news articles by extracting knowledge graphs and augmenting them through semantic clustering to recover missing links, allowing us to generate diverse and difficulty-controlled queries. Central to our framework is a 2D difficulty matrix that combines generator-side and retriever-side difficulty. Extensive experiments show that error rates strongly correlate with our difficulty measures, validating their diagnostic utility. \textsc{GRADE} enables fine-grained analysis of RAG performance and provides a scalable foundation for evaluating and improving multi-hop reasoning in real-world applications.


\end{abstract}

\section{Introduction}

Retrieval-Augmented Generation (RAG) has become a widely adopted strategy for enhancing large language models (LLMs) with external knowledge sources~\cite{li2024retrieval, su2024dragin}. By retrieving relevant documents and conditioning generation on this evidence, RAG systems aim to reduce hallucinations and improve factual accuracy in tasks such as open-domain QA, dialogue, and summarization~\cite{gao2023retrieval}. As LLMs continue to improve, the quality of generation in RAG systems has also advanced. However, the evaluation frameworks used to measure these systems have not kept pace. Most benchmarks still rely on simple factoid questions that fail to capture the reasoning complexity and retrieval difficulty faced in real-world scenarios~\cite{krishna2024fact, gabburo2024measuring}. Consequently, these datasets offer limited utility in understanding performance variation across tasks.

Current evaluation often lacks the granularity needed to assess query difficulty. Tasks are typically treated as uniformly difficult, without considering structural aspects such as the number of reasoning steps (multi-hop reasoning) or the distribution of evidence across multiple documents (multi-chunk retrieval). Moreover, many benchmarks rely on single-hop, single-chunk queries, and define difficulty using retriever-centric indicators such as lexical ambiguity or document length~\cite{gabburo2024measuring, salemi2024evaluating}. Furthermore, collapsing retrieval and reasoning onto a single axis complicates the attribution of errors to the retriever, the generator, or their interaction~\cite{lee2025mhts}. These coarse-grained approaches hinder nuanced analysis of RAG systems, making it difficult to isolate module-specific failure points—such as retrieval errors versus generation issues~\cite{barnett2024seven}.

Recent work has begun to recognize the importance of task difficulty in evaluating RAG systems. Some studies have proposed taxonomies that categorize queries based on retriever-side metrics such as retrieval recall, query ambiguity, or corpus complexity~\cite{de2024know}. These approaches acknowledge that retrieval difficulty varies across queries and have led to more nuanced analyses of retriever performance. However, such efforts remain largely focused on retrieval alone, without fully considering the downstream reasoning challenges faced by RAG models. Recent work~\cite{krishna2024fact} has highlighted the importance of evaluating RAG systems in a unified manner that accounts for both retrieval and reasoning. By comparing naïve and multi-step RAG systems using their proposed dataset, the authors demonstrate how evaluation design can directly influence our understanding of a model’s reasoning capabilities.

In particular, most prior work overlooks the structural complexity in real-world RAG scenarios. Many queries in practice require multi-hop reasoning across multiple semantically diverse sources, which poses significant challenges for retrieval and evidence integration. In practice, answering a complex question often requires synthesizing information scattered across multiple, semantically distant documents~\citep{lu2019answering, de2019question}. The cognitive and computational cost of such synthesis grows not just with the number of reasoning steps but also with the semantic dispersion of the supporting evidence. In particular, reasoning across documents from different topical clusters is typically more demanding than connecting closely related passages. For example, answering a multi-hop question like \textit{“What legal implications has the use of facial recognition technology had in European countries?”} requires synthesizing technical documents on facial recognition systems with legal texts or policy reports from EU jurisdictions.

Assessing RAG systems ability to handle real-world complexity requires benchmarks that accurately capture multi-hop reasoning—demanding both retrieval of diverse, complementary documents and their integration through coherent, step-wise inference. While recent benchmarks have begun to include multi-hop queires in RAG systems, many define “hops” merely in terms of the number of retrieved evidence chunks, rather than the logical compositionality of the reasoning required~\cite{tang2024multihop}. Such datasets may model retrieval under multi-evidence conditions, but often fall short of ensuring that the answer genuinely depends on connecting semantically distinct pieces of information in a structured reasoning chain. Instead, many tasks can be solved by stitching together co-located or loosely related facts, which risks overestimating a system’s reasoning capability~\cite{min2019compositional, zhao2023hop}. Moreover, because these datasets rarely enforce semantic heterogeneity across retrieved documents, they may underestimate the cognitive and computational demands of real-world multi-hop scenarios—where information must be integrated across topically diverse sources such as technical reports, legal documents, and policy analyses~\cite{joshi2024reaper}. As a result, although multi-hop datasets have become more prevalent, current evaluations still fall short in diagnosing the reasoning limitations of RAG systems under practical applications, highlighting the need for more fine-grained and structurally aware evaluation frameworks that jointly account for retrieval difficulty and reasoning complexity.

\begin{figure*}[t]
\centering
\includegraphics[width=\linewidth]{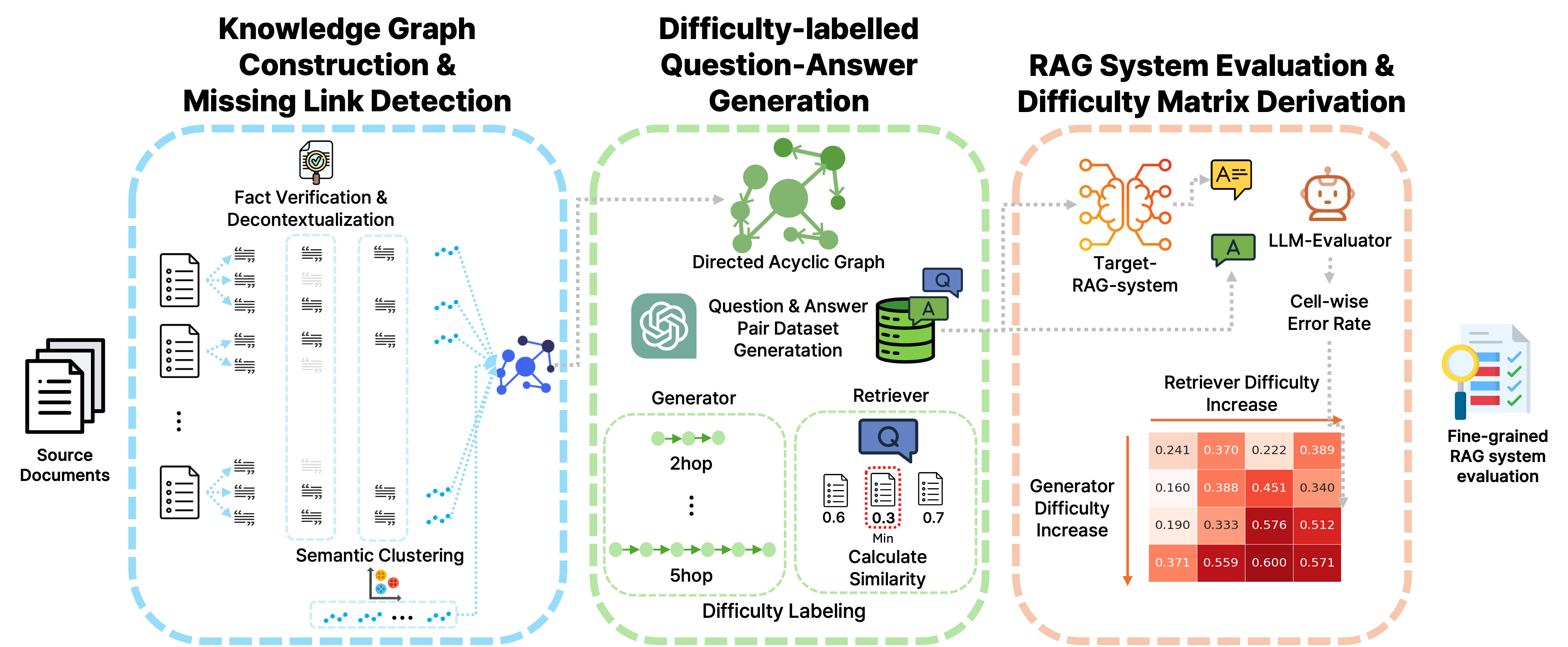}
\caption{Overall process of \textsc{GRADE} comprises three stages: (1) 
Knowledge Graph Construction \& Missing Link Detection, where factual claims are extracted from source documents, verified and decontextualized to form triples, and the KG is augmented by detecting missing links via semantic clustering;
(2) Difficulty-labelled Question–Answer Generation, where multi-hop reasoning paths in a directed acyclic graph are used to generate QA pairs, with each query assigned a difficulty label based on reasoning depth and retrieval difficulty; and (3) RAG System Evaluation \& Difficulty Matrix Derivation, where the RAG system's performance is assessed via cell-wise error rates across a 2D difficulty matrix considering both generator-side and retriever-side complexity.}
\label{fig:main_fig}
\end{figure*}

In this paper, we introduce \textsc{GRADE}, a fine-grained difficulty framework that simultaneously models two orthogonal sources of complexity: (i) multi-hop depth—the minimum number of reasoning steps required to connect the question to the answer through explicit evidence chains, and (ii) query-evidence semantic distance—the degree of semantic divergence between the query and its retrieved supporting chunks. In \textsc{GRADE}, task difficulty is defined over a two-dimensional space combining graph-based hop count and query-evidence semantic distance, where each coordinate point (i.e., the intersection of a specific hop count and semantic distance level) corresponds to a group of queries considered to share the same difficulty level. This formulation enables a fine-grained characterization of query complexity and supports (1) the synthetic generation of difficulty-calibrated test suites with controllable proportions across different difficulty levels, and (2) module-level ablations that analyze how retrieval and generation components contribute to performance at varying difficulty levels. This enables precise troubleshooting and systematic diagnosis of RAG system weaknesses.

Our main contributions are as follows: 
\begin{itemize}
    \item We propose a pipeline for multi-hop QA generation that utilizes a knowledge graph augmented through missing link detection powered by semantic clustering.
    \item We introduce a novel evaluation framework, \textsc{GRADE}, based on a difficulty matrix that jointly captures two key dimensions: reasoning depth and retrieval difficulty. This integrated approach offers a structured, fine-grained perspective on task complexity.
    \item Our framework facilitates more detailed and interpretable analysis of RAG system performance by jointly considering retrieval and generation challenges, thereby supporting more effective error diagnosis and guiding future system improvements, as illustrated in Figure~\ref{fig:main_fig}.
\end{itemize}



We release our code to facilitate reproducibility and further research.~\footnote{https://github.com/DaeyongKwon98/GRADE}

\section{Related Work}






\subsection{Limitations of Existing RAG Benchmarking Protocols}

The vast majority of RAG studies assess performance on single-hop, single-chunk factoid questions, typically taken from open-domain QA datasets such as Natural Questions, TriviaQA~\cite{lewis2020retrieval, joshi2017triviaqa}. These benchmarks were originally designed to test whether a system can retrieve a passage that explicitly contains the answer and generate it with minimal transformation. As a result, they primarily reward surface-level lexical overlap rather than capturing deeper reasoning capabilities~\cite{zheng2025reasoning, liu2023question}.

Each query in these benchmarks is evaluated using coarse metrics, such as exact match or token-level F$_1$ for generation, and hit@k or recall for retrieval, implicitly treating all queries as equally difficult. Notably, existing datasets fail to distinguish straightforward questions, whose answers directly appear within the first retrieved passage, from complex queries that require synthesizing information distributed across multiple documents. Consequently, these evaluation protocols provide limited insights into specific failure points within the RAG pipeline: poor performance might reflect retrieval failures, generation inaccuracies despite successful retrieval, or subtle discrepancies in entity normalization (e.g., recognizing “NYC” and “New York City” as the same entity)~\cite{barnett2024seven}. This conflation of retrieval and generation performance masks system weaknesses and hampers targeted improvements. Thus, more structured evaluation protocols are needed to disentangle and individually assess retrieval success and synthesis accuracy.

Recent analyses indicate that even state-of-the-art RAG systems, despite achieving near-perfect performance on conventional single-hop benchmarks, exhibit significant degradation on tasks requiring compositional reasoning or multi-document evidence integration~\cite{liu2025hoprag, khodadad2025evaluating}. These findings highlight a critical need for more granular evaluation methodologies capable of separately assessing retrieval and reasoning difficulty, rather than aggregating performance into a single coarse metric. At the same time, it is equally important to evaluate the overall complexity of the combined retrieval-generation task, as certain queries may simultaneously challenge both components~\cite{krishna2024fact}.

\subsection{Difficulty Adjustment in Evaluation Datasets}

As QA models continue to improve, it becomes crucial for evaluation datasets to evolve correspondingly in complexity to remain effective. Recent approaches to increasing evaluation difficulty often focus on controlled question generation strategies or adversarial data collection methodologies. For instance, AdversarialQA utilized a human-in-the-loop strategy, where annotators iteratively refined questions until they successfully elicited incorrect responses from models~\cite{bartolo2020beat}. This approach yielded datasets explicitly designed to surpass the capabilities of existing QA systems, thereby setting higher performance benchmarks.

Researchers have also explored dynamic approaches to proactively adjust or predict query difficulty. \textit{multHP}, a pre-retrieval predictor specifically designed to estimate the difficulty of multi-hop questions, allows for dynamic optimization of retrieval parameters such as the number of documents retrieved and supports the creation of balanced evaluation sets containing both simpler and more challenging queries \cite{samadi2023performance}. While such methods improve dataset quality and make evaluations more realistic, they often lack explicit and controllable definitions of question difficulty.  

\subsection{Multi-Hop QA Evaluation and Fine-Grained Diagnostics}
Multi-hop QA tasks require systems to synthesize information scattered across multiple documents. Compared to single-hop queries, they test a model’s ability to perform compositional reasoning, logical inference, and multi-step aggregation. To benchmark such capabilities, a series of datasets have been proposed, each aiming to increase the complexity and diagnostic power of QA evaluation.

HotpotQA introduced large-scale multi-hop questions manually constructed by crowdworkers who referenced two Wikipedia articles~\cite{yang2018hotpotqa}. HotpotQA was constructed without relying on structured triples and primarily features 2-hop questions. While sentence-level supervision enabled precise labeling, the lack of alignment with retrieval units made it less compatible with retrieval-based analysis frameworks. To address such limitations, 2WikiMultiHopQA adopted a hybrid design based on structured triples and unstructured passages, generating questions through entity overlaps across (subject, relation, object) triples~\cite{ho2020constructing}. It defined four question types—comparison, inference, compositional, and bridge comparison—with associated templates and explicitly mapped contextual support for each triple. MuSiQue later combined high-quality single-hop questions into 2–4 hop multi-hop chains using compositional templates~\cite{trivedi2022musique}, and included distractors selected via semantic similarity to promote deeper reasoning. More recently, MultiHop-RAG leveraged large language models to generate multi-hop questions and gold evidence, explicitly targeting RAG evaluation~\cite{tang2024multihop}. While these datasets advanced benchmark design, most still lack explicit control over question difficulty or retrieval complexity. In contrast, \textsc{GRADE} leverages chunk-level grounding to compute query-evidence semantic similarity directly over the corpus, enabling automatic pre-labeling of retrieval difficulty—a key prerequisite for fine-grained RAG evaluation.


Because RAG systems separate retrieval and generation modules, evaluation metrics must reflect errors at different stages of the pipeline. Conventional metrics such as exact match or F$_1$ fail to disentangle whether a wrong answer is due to retrieval failure or flawed synthesis. To overcome this, RAGChecker proposes a suite of diagnostic scores that evaluate retrieval quality and generation fidelity independently~\cite{ru2024ragchecker}. These finer-grained metrics align better with human judgments and expose trade-offs in system behavior.

\section{Methodology}

\subsection{Knowledge Graph Construction with Missing Link Detection}
\label{sec:method3_1}

To ensure that models under evaluation do not have prior exposure to the data, we collected news articles, strictly from July 2024 onward. Articles were split into sentences, and a pretrained fact/opinion classifier was used to retain only factual ones.
To enable reasoning over isolated facts, we applied a decontextualization step: each sentence was rewritten into a standalone claim whose meaning is preserved outside the original article. This was done via prompting, followed by a verification stage to ensure semantic consistency with the source sentence. 
The knowledge graph is composed of entities and relationships extracted from factual claims. Each entity is represented as a node, and each relationship connecting two entities is represented as a directed, labeled edge. A triplet $(e_i, r, e_j)$ corresponds to a $(\textit{subject}, \textit{predicate}, \textit{object})$ structure, mapping directly to a factual claim in the form of a directed relation between entities. We constructed the knowledge graph by merging entities with exact matches across triples.

However, relying solely on exact string matches is insufficient to capture all the semantically valid connections needed for multi-hop reasoning. To address this limitation and better support multi-hop traversal on the knowledge graph, we performed the following augmentation steps.
All claims were clustered using a GMM-based soft clustering approach. Within each cluster, we identified entity equivalence using an LLM to determine \textit{exact} and \textit{contextual} equivalence. For contextually equivalent entities (e.g., “the Biden administration” and “the U.S. government”) within semantically similar clusters, we augmented the knowledge graph by adding mirrored inbound and outbound edges. That is, if two entities are considered equivalent within the same cluster, any existing incoming or outgoing edges within that cluster connected to one are also added to the other. This ensures that reasoning paths remain consistent regardless of which entity representation is used. For exactly equivalent entities (e.g., “USA” and “United States”), we merged them under a canonical form.

\subsection{Question Generation}

We first enumerated all shortest directed acyclic paths of length 2 to 5 within the knowledge graph and discarded redundant paths that shared the same start and end entities while differing only in intermediate nodes. From the remaining paths, we sampled 400 per hop count to balance reasoning depth. Afterward, an additional validation step was done to remove ambiguous QA pairs, resulting in approximately 200 datas per hop. Let $P = \{(e_1, r_1, e_2), \ldots, (e_k, r_k, e_{k+1})\}$ denote a reasoning path of length $k$. We define a \textit{Logical Chaining Question (LCQ)} $q_P$ as a question for which all supporting facts are contained within the edge set $P$.



\subsection{Difficulty Matrix Construction}

\textbf{Generator-side difficulty} was quantified by grouping LCQs according to their reasoning path length $k \in {2,3,4,5}$. Since each question was generated from a $k$-hop path in the knowledge graph, we assume that higher $k$ correlates with increased compositional complexity, as answering higher-hop questions typically requires logically connecting more pieces of supporting evidence.

\noindent \textbf{Retriever-side difficulty} was defined based on the semantic distance between a question $q$ and its associated supporting chunks. Let $C_q = \{c_1, \ldots, c_n\}$ be the chunk set and $s(q, c_i)$ be the embedding similarity between $q$ and chunk $c_i$. We define the retrieval difficulty score $D_r$ as:
\begin{equation}
    D_{\text{r}}(q) = 1-\min_{c_i \in C_q} s(q, c_i)
\end{equation}

This minimum similarity serves as a bottleneck indicator: if even one required chunk is poorly retrieved, the entire reasoning chain may fail.

\noindent \textbf{Difficulty matrix} $M \in \mathbb{R}^{4 \times 4}$ is constructed, where each row corresponds to a hop count $k \in \{2,3,4,5\}$ and each column to a quartile bin of retrieval difficulty scores.

\subsection{RAG Evaluation}

To evaluate RAG performance under our framework \textsc{GRADE}, we first segmented all articles into overlapping chunks and embedded them using a frozen encoder. For each question, we retrieved the top-10 chunks and passed them to a generator model to produce an answer. Output responses were then judged by an LLM-based evaluator.

While this setup corresponds to a standard single-step RAG, we also conducted additional experiments, reported in Appendix~\ref{sec:other_rag}, with Multi-step RAG and Query-decomposition RAG. These variants address more complex information needs by enabling iterative retrieval and refinement or by decomposing questions into simpler sub-queries. Incorporating them allows us to evaluate whether more advanced retrieval strategies yield improvements beyond the baseline.

To further analyze performance across different levels of difficulty, we categorized questions using our difficulty matrix. Each cell in the matrix represents a group of queries with a combined difficulty level, jointly determined by both generator-side and retriever-side factors. For each cell, we computed the average error rate across the contained queries. The upper-left region of the matrix corresponds to lower-difficulty queries (e.g., 2-hop with low $D_r$), while the bottom-right region corresponds to higher-difficulty queries (e.g., 5-hop with high $D_r$).

\section{Experiments}

\subsection{Experimental Setup}

\begin{figure}[t]
\centering
\includegraphics[width=\linewidth]{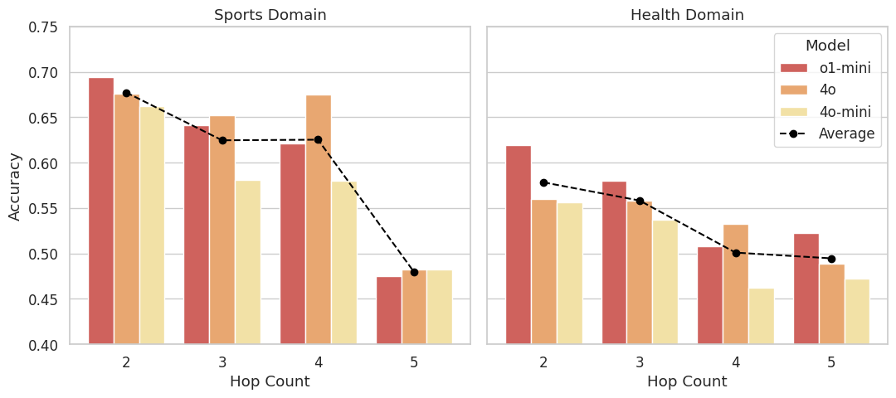}
\caption{Accuracy trends across increasing hop counts.}
\label{fig:bar_plot}
\end{figure}

Following prior work, we simulate a realistic RAG scenario in which the knowledge source diverges from the pretraining data of large language models~\cite{tang2024multihop}. To this end, we construct a news dataset using the mediastack API, a RESTful interface that aggregates global news content~\footnote{\url{https://mediastack.com/}}. The dataset comprises articles from a diverse set of English-language sources spanning domains such as sports, health, and science. We focus on articles published between July 2024 and April 2025—a period beyond the knowledge cutoff dates of the LLMs we used—thereby minimizing the likelihood that the base models have seen this content during training. Articles with fewer than 512 tokens were excluded to ensure a minimum level of contextual complexity, while those exceeding 8,192 tokens were removed to avoid excessively large inputs that could hinder processing efficiency. For the RAG database, we split each article into chunks with a total length of 128 to 256 tokens, using a 50-token overlap between consecutive chunks. We embedded them using four representative embedding models: OpenAI~\textit{text-embedding-3-small}, \textit{all-mpnet-base-v2}, \textit{bge-en-large-v1.5}, and \textit{jina-embeddings-v3}. This selection covers both closed-source and open-source approaches, ensuring a balanced and comprehensive evaluation across different embedding paradigms.

To evaluate performance across a spectrum of capabilities, we select five representative LLMs: GPT-4o, GPT-4o mini, o1-mini, Claude-4-Sonnet, and Llama 3.2 3B Instruct~\footnote{\url{https://ai.meta.com/blog/llama-3-2-connect-2024-vision-edge-mobile-devices/}}. 
This selection enables a balanced comparison across general-purpose, lightweight, reasoning-focused, and open-source LLMs. All prompts are provided in the appendix~\ref{app:prompt-design} for transparency and reproducibility.

\begin{table}[t]
\centering
\resizebox{\linewidth}{!}{%
\begin{tabular}{@{}cccccccc@{}}
\toprule
\multirow{2}{*}{\textbf{Hop}} & \multicolumn{3}{c}{\textbf{Sports}} & \multicolumn{1}{l}{} & \multicolumn{3}{c}{\textbf{Health}} \\
                     & \textbf{o1-mini}  & \textbf{4o}    & \textbf{4o-mini} &            & \textbf{o1-mini}  & \textbf{4o}    & \textbf{4o-mini} \\ \midrule
2-hop                & 0.504    & 0.463 & 0.646   &                      & 0.967    & 0.671 & 0.806   \\
3-hop                & 0.595    & 0.483 & 0.433   &                      & 0.699    & 0.885 & 0.956   \\
4-hop                & 0.976    & 0.963 & 0.769   &                      & 0.950     & 0.874 & 0.802   \\
5-hop                & 0.867    & 0.798 & 0.664   &                      & 0.704    & 0.838 & 0.968   \\ \midrule
Average              & 0.736    & 0.677 & 0.628   &                      & 0.830    & 0.817 & 0.883   \\ \bottomrule
\end{tabular}
}
\caption{\textbf{Pearson correlation} between retrieval difficulty $D_r$ and error rate, computed separately for each fixed hop count, across domains and models.}
\label{tab:retriever_correlation}
\end{table}

\subsection{Main Results}
\label{4_2_main_results}


To validate the effectiveness of our proposed difficulty framework, we evaluated generator-side, retriever-side, and combined difficulty matrix.  We first assessed generator-side difficulty by examining whether increasing the number of hops corresponds to more challenging multi-hop questions. Specifically, we compared the average accuracy across hop counts ranging from 2 to 5. As shown in Figure~\ref{fig:bar_plot}, both the Sports and Health domains exhibited a consistent decrease in average accuracy as the number of hops increased. For example, in the Sports domain, accuracy dropped from 0.68 at 2-hop to 0.48 at 5-hop—a 20\% decrease. In the Health domain, accuracy declined from 0.58 to 0.49 over the same hop range, showing a 9\% drop. This supports our hypothesis that generator-side difficulty increases with hop count, as answering higher-hop questions typically requires logically connecting more pieces of supporting evidence.

\begin{figure*}[t]
\centering
\includegraphics[width=\linewidth]{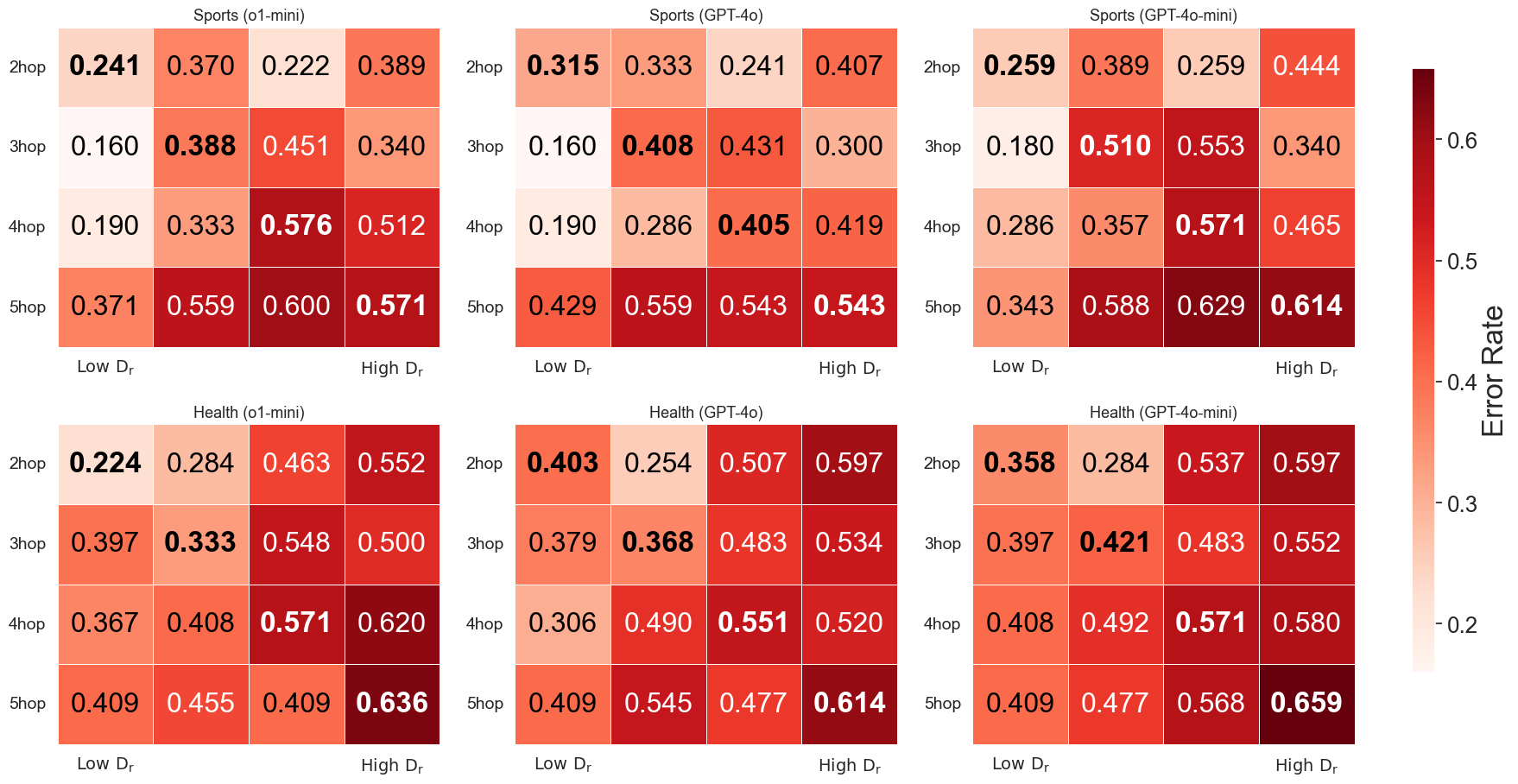}
\caption{Error rates (1-accuracy) across varying levels of reasoning depth (2–5 hops) and retrieval difficulty (low vs. high) in both Health and Sports domains.}
\label{fig:main_results_fig}
\end{figure*}

To evaluate retriever-side difficulty, we controlled for generator-side factors by fixing the hop count, ensuring that variations in performance were attributable solely to the retriever. For each fixed hop level, we split the data into four bins based on the retrieval difficulty score $D_r$, as defined earlier, and computed the average accuracy for each bin. We then calculated the Pearson correlation between retrieval difficulty and error rate for each hop level. As shown in Table~\ref{tab:retriever_correlation}, all models exhibited a strong positive correlation, averaging above 0.6 in the Sports domain and above 0.8 in the Health domain. This indicates that the minimum similarity between the query and its most weakly aligned supporting chunk increases retrieval difficulty $D_r$, thereby leading to reduced accuracy.

Finally, to evaluate the combined difficulty framework \textsc{GRADE} that considers both generator-side and retriever-side difficulty, we constructed a $4 \times 4$ difficulty matrix where the vertical axis corresponds to reasoning depth (2–5 hops) and the horizontal axis corresponds to retrieval difficulty (divided into four bins based on $D_r$ scores). For each cell in the matrix, we computed the average error rate and visualized the results in Figure~\ref{fig:main_results_fig}.

\begin{table}[t]
\centering
\resizebox{\linewidth}{!}{%
\begin{tabular}{@{}lcccccc@{}}
\toprule
\multirow{2}{*}{\textbf{Hop}} & \multicolumn{2}{c}{\textbf{Exact Eql. (E)}} & \multicolumn{2}{c}{\textbf{Context Eql. (C)}} & \multicolumn{2}{c}{\textbf{Overall (E+C)}} \\
                     & \textbf{Health}         & \textbf{Sports}         & \textbf{Health}          & \textbf{Sports}          & \textbf{Health}          & \textbf{Sports}          \\ \midrule
2-hop                & 45.9           & 10.7           & 4.85            & 9.26            & 50.8            & 19.9            \\
3-hop                & 46.3           & 11.6           & 5.19            & 8.08            & 51.5            & 19.7            \\
4-hop                & 57.4           & 23.7           & 3.05            & 10.7            & 60.4            & 34.3            \\
5-hop                & 74.4           & 29.5           & 0.00               & 7.91            & 74.4            & 37.4            \\ \bottomrule
\end{tabular}
}
\caption{Percentage of queries that require semantic clustering to resolve missing links, broken down by hop count and domain.}
\label{tab:hop-performance}
\vspace{-1em}
\end{table}

Across all models and in both the Sports and Health domains, we observed a general trend: error rates tended to be lower in the upper-left region of the matrix (2-hop, low $D_r$) and higher in the bottom-right region (5-hop, high $D_r$), with performance degrading steadily along both dimensions in most cases. Notably, values along the diagonal from upper-left to bottom-right showed the steepest increase in error rate, indicating that tasks involving both deeper reasoning and harder retrieval are significantly more difficult. All matrices exhibited strong linearity across the four diagonal points, with Pearson correlation coefficients exceeding 0.9—highlighting the consistency of this trend. For example, in the Sports domain with o1-mini, the error rate increased from 0.241 in the upper-left cell (2-hop, low $D_r$) to 0.571 in the bottom-right cell (5-hop, high $D_r$), marking a 33\% increase. In the Health domain, the difference was even more pronounced, with error rates rising from 0.224 to 0.636—a 41.2\% increase. While the overall error rates were higher in Health than in Sports, the gap between easy and hard questions was more prominent in the Sports domain. 
These results support the validity of our difficulty framework, demonstrating that jointly modeling both generator-side and retriever-side challenges yields an accurate, fine-grained, and diagnostic approach to evaluating multi-hop QA performance in RAG systems.
The results of our experiments with additional models, embeddings, RAG systems, domains, and chunk sizes are reported in the Appendix~\ref{sec:appendix}.

\subsection{Analysis on Missing Links}

In traditional triplet-based multi-hop QA generation pipelines, triplets are connected using shared entities that require an exact match, which often leads to missing links. For example, different expressions referring to the same entity, such as \textit{"USA"} and \textit{"United States,"} need to be identified and linked. Similarly, cases where phrases like \textit{"Biden administration"} and \textit{"U.S. government"} refer to the same entity within a specific context also require connection. These missing links represent semantically similar but not explicitly connected entities.
To address this, we developed a method described in Section~\ref{sec:method3_1} that uses semantic clustering and LLMs to supplement the knowledge graph with two types of missing links: \textit{exact} and \textit{contextually} equivalent entities. The proportion of data augmented with these missing links, broken down by hop count and domain, is presented in Table~\ref{tab:qual-missing-links}.


We observe that the impact of missing links becomes increasingly pronounced as the number of reasoning hops grows. In the Health domain, the proportion of data augmented with missing links increased from 50.8\% at 2-hop to 74.4\% at 5-hop. In the Sports domain, it rose from 19.9\% to 37.4\%. The Health data showed a particularly high rate of exact equivalents and a relatively lower rate of contextually equivalent links. This is likely due to the frequent occurrence of common entities such as drugs and diseases in the Health domain. Although the proportions vary depending on the domain characteristics, these results demonstrate that a significant portion of multi-hop data benefits from the inclusion of missing links.

\subsection{Ablation Study}

\begin{table}[]
\centering
\resizebox{\columnwidth}{!}{
\begin{tabular}{@{}clccc@{}}
\toprule
\textbf{Domain}                  & \textbf{Agg. Func.}     & \textbf{o1-mini}        & \textbf{4o}             & \textbf{4o-mini}        \\ \midrule
\multirow{3}{*}{Sports} & Mean       & 0.445          & 0.394          & 0.461          \\
                        & Power Mean & 0.534          & 0.398          & 0.498          \\
                        & Min        & \textbf{0.736} & \textbf{0.677} & \textbf{0.628} \\ \midrule
\multirow{3}{*}{Health} & Mean       & 0.314          & 0.412          & 0.416          \\
                        & Power Mean & \textbf{0.903} & 0.795          & 0.808          \\
                        & Min        & 0.830           & \textbf{0.817} & \textbf{0.883} \\ \bottomrule
\end{tabular}
}
\caption{\textbf{Pearson correlation} between different aggregation functions of retrieval difficulty $D_r$ and error rate, across domains and models.}
\label{tab:ablation}
\end{table}

To verify that the retrieval-difficulty score used in \textsc{GRADE} faithfully tracks task complexity, we ablated the aggregation function that collapses the set of query–chunk embedding similarities into a single scalar. For every question we computed (i) the arithmetic Mean, (ii) a power mean with exponent $p=-2$ (emphasizing low-similarity entries while retaining signal from the rest), and (iii) the Minimum similarity. We then measured the Pearson correlation between each variant and error rate while holding hop depth constant. As Table~\ref{tab:ablation} shows, Min delivers the strongest alignment with error rate in the Sports domain surpassing the arithmetic mean by roughly 29\% and the power mean by 16\% on average.

Across the six model–domain pairs, the Min aggregation delivers the highest (or co-highest) Pearson correlation with error rate in four cases and never falls below second place (Table 2). This regularity suggests that overall task difficulty is usually governed by the single hardest retrieval hop: when any required chunk is missed, the reasoning chain cannot be completed and the question is likely answered incorrectly. Because the minimum similarity directly captures this bottleneck, it provides a upper bound on achievable performance.

The power mean ($p=-2$) achieves a slightly lower correlation than the minimum, but consistently outperforms the arithmetic mean in all settings. The power mean smooths over noisy signals by down-weighting over-retrieved chunks, its benefit may be particularly visible in domains like Health, where essential information is expressed across multiple documents using diverse surface forms.  Nonetheless, since Logical Chaining Questions (LCQ) require all intermediate chunks to be retrieved for correct inference, the minimum similarity remains the most reliable indicator of retrieval-side difficulty and is adopted as our default metric in all subsequent analyses.

\section{Conclusion}


We present GRADE, a novel evaluation framework for Retrieval-Augmented Generation (RAG) that jointly considers reasoning depth (generator-side) and semantic distance (retriever-side) to better reflect the structural complexity of real-world multi-hop question answering. By defining a difficulty matrix based on hop count and retrieval difficulty, our approach enables fine-grained analysis of RAG systems, disentangling the contributions of retrieval and generation. Experimental results show that our difficulty metrics strongly correlate with error rates across multiple RAG models in realistic settings. In addition, our knowledge graph augmentation—through missing link detection via semantic clustering—highlights the limitations of traditional triplet-based pipelines in supporting deeper reasoning. Notably, the GRADE difficulty matrix itself serves as a unified difficulty space, where diagonal regions represent scenarios jointly challenging for both retrieval and generation. This holistic perspective enables more nuanced diagnoses of RAG system behavior across the full complexity spectrum. Overall, \textsc{GRADE} offers a scalable, interpretable tool for analyzing RAG system failures and guiding future advancements in multi-hop reasoning architectures.

Future work will extend \textsc{GRADE} to enable module-level evaluation across the broader RAG pipeline, including components such as query decomposition and document filtering.

\newpage

\section*{Limitations}

\paragraph{LLM Dependency in Pipeline Components}
Several core components of \textsc{GRADE}—including claim decontextualization, entity equivalence detection (both exact and contextual), and answer evaluation—depend heavily on large language models (LLMs). While this allows for scalable and flexible data generation, it introduces potential sources of noise and bias. LLMs may hallucinate facts during claim rewriting, inconsistently judge contextual equivalence between entities, or misclassify correct answers due to surface-level mismatches in automatic evaluation. These failure modes can propagate through the pipeline, ultimately affecting the fidelity of the synthetic dataset and the reliability of difficulty estimation. Future work may benefit from integrating more robust validation mechanisms, such as human-in-the-loop verification or hybrid approaches combining symbolic and statistical techniques.

\paragraph{Dimensionality of the Difficulty Matrix}
Our proposed difficulty matrix models task complexity along two axes: reasoning depth and retrieval difficulty, corresponding to generator- and retriever-side challenges. However, real-world RAG systems often involve additional modules such as query decomposition, routing, or document filtering, which also significantly influence task difficulty. Incorporating these components would require extending the difficulty space into a higher-dimensional representation, potentially enabling a richer understanding of how different system modules interact under compositional reasoning.

\paragraph{Limited Scope of Reasoning Types}
In this work, we restrict our focus to reasoning types where the correct answer is explicitly entailed by the content of the retrieved documents. That is, each query is designed such that the answer can be derived directly through logical chaining of facts present within the corpus, without requiring speculative or implicit inference. While this approach ensures controllability and verifiability of reasoning paths, it excludes more complex or open-ended reasoning types—such as analogical, causal, or commonsense inference—that are not strictly grounded in the retrieved evidence. Expanding the \textsc{GRADE} framework to support such reasoning paradigms remains a compelling direction for future work.

\section*{Ethics Statements}
Our framework leverages large language models (LLMs) in various stages, including claim decontextualization, entity equivalence detection, and automatic evaluation. While LLMs provide scalability and flexibility, they are also known to reflect societal biases or produce harmful content, including discriminatory or offensive expressions~\cite{gallegos2024bias}. As such, our method may inadvertently introduce such risks into the generated dataset or evaluation outputs. Although we attempt to mitigate this by sourcing factual news content and excluding opinionated material, residual bias or inappropriate generation remains a concern. We encourage future work to incorporate stronger content filtering, human oversight, and bias mitigation strategies, particularly in downstream applications.




\appendix

\clearpage

\section{Appendix}
\label{sec:appendix}

\subsection{Additional Main Results}

To verify the generalizability of our proposed difficulty framework, we replicated the main experiments using a science-focused corpus. As in the Sports and Health domains, we selected factual science articles published after July 2024 and constructed a new knowledge graph following the methodology described in Section~\ref{sec:method3_1}.

\begin{figure*}[ht]
\centering
\includegraphics[width=\linewidth]{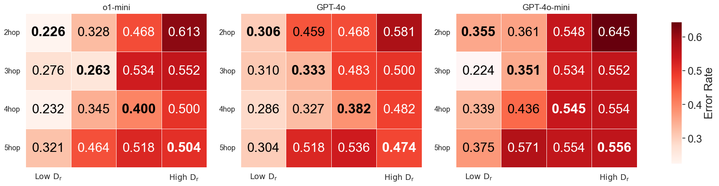}
\caption{Model performance across varying levels of reasoning depth (2–5 hops) and retrieval difficulty (low vs. high) in \textbf{Science} domain. Each heatmap cell represents the error rate for a given setting.}
\label{fig:science_matrix}
\end{figure*}

Figure~\ref{fig:science_matrix} presents the 2D difficulty matrices for three models—o1-mini, GPT-4o, and GPT-4o-mini—evaluated on the science corpus. Across all models, the diagonal entries (i.e., cells where hop count and retrieval difficulty increase together) exhibit a clear upward trend in error rate. For example, o1-mini shows error rates of 0.226 (2-hop, low $D_r$), rising steadily to 0.504 (5-hop, high $D_r$). This pattern is mirrored in GPT-4o (0.306 to 0.474) and GPT-4o-mini (0.355 to 0.556), reaffirming that our difficulty framework captures compounding complexity from both axes. These diagonals provide strong empirical support for the framework’s discriminative power.

Overall, the diagonal consistency observed across all three models underscores the robustness of our difficulty framework, while deviations from monotonicity in off-diagonal cells point to complex interactions between retrieval and reasoning. These findings highlight the utility of our 2D framework not only as a ranking tool for task difficulty, but also as a diagnostic surface for probing nuanced model behaviors under hybrid conditions.

\subsection{Qualitative Results}

While conventional symbolic pipelines rely on explicit triplet connections, many real-world reasoning chains involve latent semantic associations not captured by direct edges. To address this, we incorporate entity equivalence—both exact and contextual—via semantic clustering, enabling reasoning over missing links that would otherwise be invisible to standard multi-hop traversal.

Table~\ref{tab:qual-missing-links} illustrates two representative cases. In the first example, two entities (\textit{“elevated ketamine levels in his blood”} and \textit{“elevated ketamine levels”}) are treated as contextually equivalent within the same semantic cluster, allowing the model to infer \textit{“respiratory depression”} as the consequence of ketamine exposure. The second example highlights exact equivalence, where mentions like \textit{“bird flu (H5N1)”} and \textit{“H5N1 virus”} refer to the same concept, enabling the system to reason across mentions and infer a reduction in lethality. These examples demonstrate the necessity of enriched graph construction for supporting robust multi-hop reasoning.

\begin{table}[t]
\centering
\small
\begin{tabular}{p{0.95\linewidth}}
\toprule
\textbf{Contextually Same Example} \\
\midrule
\textbf{Triples:} \\
(matthew perry, had, \textcolor{red}{elevated ketamine levels in his blood}) \\
(\textcolor{red}{elevated ketamine levels}, led to, respiratory depression) \\
\textbf{Question:} What condition did actor Matthew Perry experience due to elevated ketamine levels in his blood? \\
\textbf{Answer:} Respiratory depression \\
\midrule
\textbf{Exact Same Example} \\
\midrule
\textbf{Triples:} \\
(human spillover cases, will persist, \textcolor{blue}{bird flu (h5n1)}) \\
(\textcolor{blue}{h5n1 virus}, indicates, potential reduction in its lethality) \\
\textbf{Question:} What does the persistence of human spillover cases of bird flu (H5N1) indicate about bird flu's lethality? \\
\textbf{Answer:} Potential reduction in its lethality \\
\bottomrule
\end{tabular}
\caption{Qualitative examples of missing link reasoning using contextually and exactly equivalent entities.}
\label{tab:qual-missing-links}
\end{table}

\subsection{Additional Model Families}

We extend the evaluation to two non-OpenAI models---\textbf{Claude 4 Sonnet (20250514)} and \textbf{Llama 3.2 3B Instruct}.
Both were tested on three representative domains (Sports, Health, Science) under the same setup as described in the main experiment. 
Across all cases, error rates increase consistently with both hop count and retrieval difficulty (\(D_r\)).
This confirms that the proposed difficulty formulation produces stable trends across different model families.

\begin{table*}[t]
\centering
\small
\begin{tabular}{llcccc}
\toprule
Domain & Hops & $Q_1$ (Low $D_r$) & $Q_2$ & $Q_3$ & $Q_4$ (High $D_r$) \\
\midrule
\multirow{4}{*}{Sports}
 & 2-hop & \textbf{0.222} & 0.333 & 0.241 & 0.407 \\
 & 3-hop & 0.160 & \textbf{0.388} & 0.592 & 0.460 \\
 & 4-hop & 0.262 & 0.310 & \textbf{0.429} & 0.535 \\
 & 5-hop & 0.400 & 0.529 & 0.600 & \textbf{0.457} \\
\midrule
\multirow{4}{*}{Health}
 & 2-hop & \textbf{0.343} & 0.299 & 0.522 & 0.716 \\
 & 3-hop & 0.414 & \textbf{0.491} & 0.534 & 0.724 \\
 & 4-hop & 0.408 & 0.531 & \textbf{0.633} & 0.680 \\
 & 5-hop & 0.477 & 0.591 & 0.659 & \textbf{0.795} \\
\midrule
\multirow{4}{*}{Science}
 & 2-hop & \textbf{0.262} & 0.350 & 0.525 & 0.689 \\
 & 3-hop & 0.300 & \textbf{0.450} & 0.517 & 0.656 \\
 & 4-hop & 0.246 & 0.526 & \textbf{0.456} & 0.509 \\
 & 5-hop & 0.352 & 0.453 & 0.648 & \textbf{0.537} \\
\bottomrule
\end{tabular}
\caption{Error rates for Claude 4 Sonnet (20250514).}
\label{tab:claude-appendix}
\end{table*}

\begin{table*}[t]
\centering
\small
\begin{tabular}{llcccc}
\toprule
Domain & Hops & $Q_1$ (Low $D_r$) & $Q_2$ & $Q_3$ & $Q_4$ (High $D_r$) \\
\midrule
\multirow{4}{*}{Sports}
 & 2-hop & \textbf{0.519} & 0.556 & 0.463 & 0.630 \\
 & 3-hop & 0.420 & \textbf{0.633} & 0.796 & 0.720 \\
 & 4-hop & 0.452 & 0.571 & \textbf{0.786} & 0.860 \\
 & 5-hop & 0.514 & 0.735 & 0.943 & \textbf{0.686} \\
\midrule
\multirow{4}{*}{Health}
 & 2-hop & \textbf{0.262} & 0.350 & 0.525 & 0.689 \\
 & 3-hop & 0.300 & \textbf{0.450} & 0.517 & 0.656 \\
 & 4-hop & 0.246 & 0.526 & \textbf{0.456} & 0.509 \\
 & 5-hop & 0.352 & 0.453 & 0.648 & \textbf{0.537} \\
\midrule
\multirow{4}{*}{Science}
 & 2-hop & \textbf{0.508} & 0.700 & 0.803 & 0.836 \\
 & 3-hop & 0.533 & \textbf{0.633} & 0.650 & 0.738 \\
 & 4-hop & 0.561 & 0.702 & \textbf{0.649} & 0.702 \\
 & 5-hop & 0.537 & 0.698 & 0.815 & \textbf{0.685} \\
\bottomrule
\end{tabular}
\caption{Error rates for Llama 3.2 3B Instruct.}
\label{tab:llama-appendix}
\end{table*}

\subsection{Evaluation on Multi-Step and Decomposition RAG Systems}\label{sec:other_rag}

We extend the evaluation to two stronger RAG pipelines under the same protocol and difficulty matrix as in the main paper. 
(1) \textbf{Multi-step RAG}: two-stage retrieval and generation—first generate an intermediate answer from top-10 retrieved documents, then perform a second retrieval conditioned on the original query, retrieved context, and the intermediate answer to produce the final response. 
(2) \textbf{Query-decomposition RAG}: decompose the query into sub-queries; for each sub-query retrieve top-10 documents, merge with de-duplication, and rerank using \texttt{bge-reranker-large}. 
All results are reported as cell-wise error rates \((1-\text{accuracy})\) over the 4$\times$4 difficulty matrix (rows = hop count; columns = \(D_r\) quartiles from low to high).

In both systems and across domains, error rates generally increase with hop count and retrieval difficulty, with the largest gains along the main diagonal of the matrix (higher hops and higher $D_r$).

\subsection{Evaluation on Entertainment and Technology Domains}

To further examine the robustness of our framework across domains, we extended the experiments to Entertainment and Technology corpora. Following the same pipeline described in Section~3, we constructed domain-specific knowledge graphs and generated multi-hop queries (2--5 hops). For both domains, error rates showed consistent trends with our main analysis: performance degrades as reasoning depth increases and retrieval difficulty grows, while overall tendencies align with the Sports, Health, and Science results.

As shown in Table~\ref{tab:app_ent} and Table~\ref{tab:app_tech}, these additional experiments reinforce the generalizability of the proposed difficulty framework across diverse topical domains. The observed patterns mirror those in the primary analysis, further supporting the consistency of GRADE under varying domain conditions.
\subsection{Human Evaluation on Missing Link Detection}
To assess the reliability of our missing link detection process, we manually analyzed 30 randomly sampled entity groups from the augmentation phase.
We observed 5 ambiguous cases among exact-equivalence groups and 6 among contextually-equivalent groups. Most were not outright errors but borderline situations involving partial overlaps or differences in granularity. For example, \textit{“the first round of the draft”} versus \textit{“the first round of the NFL draft”} reflected an inclusion rather than strict synonymy, while \textit{“redemption”} versus \textit{“Michael Vick’s ongoing redemption”} showed a mismatch in abstraction level.
Such cases were limited in scope relative to the overall mappings and did not noticeably affect question generation or evaluation outcomes. Still, we note that refining entity matching strategies remains a worthwhile direction, especially in high-stakes domains.





\begin{table*}[h]
\centering
\small
\begin{tabular}{llcccc}
\toprule
\textbf{Multi-step RAG} & \textbf{Hops} & \textbf{$Q_1$ (Low $D_r$)} & \textbf{$Q_2$} & \textbf{$Q_3$} & \textbf{$Q_4$ (High $D_r$)} \\
\midrule
\multirow{4}{*}{Sports}
 & 2-hop & \textbf{0.352} & 0.296 & 0.185 & 0.444 \\
 & 3-hop & 0.180 & \textbf{0.388} & 0.490 & 0.420 \\
 & 4-hop & 0.357 & 0.333 & \textbf{0.476} & 0.465 \\
 & 5-hop & 0.371 & 0.706 & 0.600 & \textbf{0.457} \\
\midrule
\multirow{4}{*}{Health}
 & 2-hop & \textbf{0.403} & 0.299 & 0.507 & 0.672 \\
 & 3-hop & 0.310 & \textbf{0.421} & 0.466 & 0.569 \\
 & 4-hop & 0.388 & 0.490 & \textbf{0.592} & 0.660 \\
 & 5-hop & 0.591 & 0.500 & 0.568 & \textbf{0.727} \\
\midrule
\multirow{4}{*}{Science}
 & 2-hop & \textbf{0.262} & 0.383 & 0.557 & 0.721 \\
 & 3-hop & 0.367 & \textbf{0.483} & 0.550 & 0.705 \\
 & 4-hop & 0.316 & 0.491 & \textbf{0.439} & 0.509 \\
 & 5-hop & 0.352 & 0.472 & 0.630 & \textbf{0.537} \\
\bottomrule
\end{tabular}
\caption{Extended evaluation on a two-stage \textbf{Multi-step RAG}. Values are error rates over the 4$\times$4 difficulty matrix (rows = hop count; columns = $D_r$ quartiles). Lower is better.}
\label{tab:msrag}
\end{table*}

\begin{table*}[h]
\centering
\small
\begin{tabular}{llcccc}
\toprule
\textbf{Query-decomposition RAG} & \textbf{Hops} & \textbf{$Q_1$ (Low $D_r$)} & \textbf{$Q_2$} & \textbf{$Q_3$} & \textbf{$Q_4$ (High $D_r$)} \\
\midrule
\multirow{4}{*}{Sports}
 & 2-hop & \textbf{0.519} & 0.500 & 0.426 & 0.500 \\
 & 3-hop & 0.360 & \textbf{0.531} & 0.653 & 0.580 \\
 & 4-hop & 0.500 & 0.738 & \textbf{0.738} & 0.605 \\
 & 5-hop & 0.514 & 0.588 & 0.629 & \textbf{0.743} \\
\midrule
\multirow{4}{*}{Health}
 & 2-hop & \textbf{0.478} & 0.493 & 0.627 & 0.746 \\
 & 3-hop & 0.690 & \textbf{0.754} & 0.776 & 0.672 \\
 & 4-hop & 0.408 & 0.714 & \textbf{0.714} & 0.740 \\
 & 5-hop & 0.705 & 0.750 & 0.750 & \textbf{0.773} \\
\midrule
\multirow{4}{*}{Science}
 & 2-hop & \textbf{0.443} & 0.500 & 0.623 & 0.787 \\
 & 3-hop & 0.550 & \textbf{0.533} & 0.717 & 0.705 \\
 & 4-hop & 0.491 & 0.684 & \textbf{0.579} & 0.649 \\
 & 5-hop & 0.648 & 0.528 & 0.630 & \textbf{0.741} \\
\bottomrule
\end{tabular}
\caption{Extended evaluation on a \textbf{Query-decomposition RAG} with per-subquery retrieval, merge \& de-duplication, and \texttt{bge-reranker-large} reranking. Values are error rates.}
\label{tab:qdrag}
\end{table*}

\begin{table}[h]
\centering
\begin{tabular}{c|cccc}
\toprule
Hops & Low $D_r$ & Q2 & Q3 & High $D_r$ \\
\midrule
2-hop & \textbf{0.375} & 0.298 & 0.489 & 0.479 \\
3-hop & 0.471 & \textbf{0.440} & 0.549 & 0.608 \\
4-hop & 0.378 & 0.523 & \textbf{0.689} & 0.578 \\
5-hop & 0.708 & 0.522 & 0.652 & \textbf{0.542} \\
\bottomrule
\end{tabular}
\caption{Error rates in the \textsc{Entertainment} domain. Lower is better.}
\label{tab:app_ent}
\end{table}

\begin{table}[h]
\centering
\begin{tabular}{c|cccc}
\toprule
Hops & Low $D_r$ & Q2 & Q3 & High $D_r$ \\
\midrule
2-hop & \textbf{0.431} & 0.380 & 0.420 & 0.608 \\
3-hop & 0.417 & \textbf{0.375} & 0.542 & 0.667 \\
4-hop & 0.386 & 0.465 & \textbf{0.488} & 0.545 \\
5-hop & 0.564 & 0.564 & 0.487 & \textbf{0.550} \\
\bottomrule
\end{tabular}
\caption{Error rates in the \textsc{Technology} domain. Lower is better.}
\label{tab:app_tech}
\end{table}

\subsection{Evaluation with Alternative Embedding Models}

\begin{table}[h!]
\centering
\small
\begin{tabular}{lcccc}
\toprule
\textbf{Hops} & \textbf{Low $D_r$} & Q2 & Q3 & \textbf{High $D_r$} \\
\midrule
\multicolumn{5}{c}{\texttt{all-mpnet-base-v2}} \\
2-hop & \textbf{0.296} & 0.121 & 0.241 & 0.389 \\
3-hop & 0.220 & \textbf{0.388} & 0.469 & 0.360 \\
4-hop & 0.262 & 0.262 & \textbf{0.452} & 0.442 \\
5-hop & 0.286 & 0.441 & 0.571 & \textbf{0.514} \\
\midrule
\multicolumn{5}{c}{\texttt{bge-en-large-v1.5}} \\
2-hop & \textbf{0.315} & 0.111 & 0.222 & 0.333 \\
3-hop & 0.180 & \textbf{0.286} & 0.449 & 0.400 \\
4-hop & 0.191 & 0.310 & \textbf{0.405} & 0.442 \\
5-hop & 0.371 & 0.265 & 0.429 & \textbf{0.686} \\
\midrule
\multicolumn{5}{c}{\texttt{jina-embeddings-v3}} \\
2-hop & \textbf{0.315} & 0.167 & 0.259 & 0.444 \\
3-hop & 0.140 & \textbf{0.388} & 0.469 & 0.600 \\
4-hop & 0.167 & 0.357 & \textbf{0.595} & 0.651 \\
5-hop & 0.400 & 0.353 & 0.543 & \textbf{0.771} \\
\bottomrule
\end{tabular}
\caption{Error rates in the \textsc{Sports} domain with three embedding models.}
\label{tab:emb_sports}
\end{table}

\begin{table}[h!]
\centering
\small
\begin{tabular}{lcccc}
\toprule
\textbf{Hops} & \textbf{Low $D_r$} & Q2 & Q3 & \textbf{High $D_r$} \\
\midrule
\multicolumn{5}{c}{\texttt{all-mpnet-base-v2}} \\
2-hop & \textbf{0.343} & 0.254 & 0.388 & 0.731 \\
3-hop & 0.310 & \textbf{0.421} & 0.552 & 0.603 \\
4-hop & 0.347 & 0.531 & \textbf{0.510} & 0.600 \\
5-hop & 0.386 & 0.659 & 0.477 & \textbf{0.545} \\
\midrule
\multicolumn{5}{c}{\texttt{bge-en-large-v1.5}} \\
2-hop & \textbf{0.299} & 0.299 & 0.388 & 0.612 \\
3-hop & 0.362 & \textbf{0.404} & 0.483 & 0.448 \\
4-hop & 0.347 & 0.531 & \textbf{0.633} & 0.460 \\
5-hop & 0.500 & 0.523 & 0.682 & \textbf{0.477} \\
\midrule
\multicolumn{5}{c}{\texttt{jina-embeddings-v3}} \\
2-hop & \textbf{0.313} & 0.269 & 0.463 & 0.716 \\
3-hop & 0.345 & \textbf{0.404} & 0.431 & 0.655 \\
4-hop & 0.306 & 0.571 & \textbf{0.531} & 0.560 \\
5-hop & 0.432 & 0.682 & 0.568 & \textbf{0.636} \\
\bottomrule
\end{tabular}
\caption{Error rates in the \textsc{Health} domain for three emedding models.}
\label{tab:emb_health}
\end{table}

\begin{table}[h!]
\centering
\small
\begin{tabular}{lcccc}
\toprule
\textbf{Hops} & \textbf{Low $D_r$} & Q2 & Q3 & \textbf{High $D_r$} \\
\midrule
\multicolumn{5}{c}{\texttt{all-mpnet-base-v2}} \\
2-hop & \textbf{0.230} & 0.367 & 0.492 & 0.689 \\
3-hop & 0.383 & \textbf{0.383} & 0.633 & 0.574 \\
4-hop & 0.228 & 0.474 & \textbf{0.456} & 0.544 \\
5-hop & 0.333 & 0.547 & 0.519 & \textbf{0.537} \\
\midrule
\multicolumn{5}{c}{\texttt{bge-en-large-v1.5}} \\
2-hop & \textbf{0.295} & 0.367 & 0.393 & 0.705 \\
3-hop & 0.333 & \textbf{0.500} & 0.400 & 0.623 \\
4-hop & 0.263 & 0.386 & \textbf{0.404} & 0.404 \\
5-hop & 0.352 & 0.358 & 0.500 & \textbf{0.481} \\
\midrule
\multicolumn{5}{c}{\texttt{jina-embeddings-v3}} \\
2-hop & \textbf{0.213} & 0.333 & 0.443 & 0.607 \\
3-hop & 0.367 & \textbf{0.500} & 0.433 & 0.639 \\
4-hop & 0.246 & 0.404 & \textbf{0.439} & 0.544 \\
5-hop & 0.296 & 0.453 & 0.481 & \textbf{0.611} \\
\bottomrule
\end{tabular}
\caption{Error rates in the \textsc{Science} domain for three embedding models.}
\label{tab:emb_science}
\end{table}

To further assess the robustness of the GRADE framework, 
we conducted additional experiments using three widely adopted embedding encoders: 
\texttt{all-mpnet-base-v2}~\footnote{\url{https://huggingface.co/sentence-transformers/all-mpnet-base-v2}}, 
\texttt{bge-en-large-v1.5}~\footnote{\url{https://huggingface.co/BAAI/bge-large-en-v1.5}}, 
and \texttt{jina-embeddings-v3}~\footnote{\url{https://huggingface.co/jinaai/jina-embeddings-v3}}. 
These models differ substantially in architecture, training data, and design philosophy, 
providing complementary perspectives on sentence-level representation.

As shown in Tables~\ref{tab:emb_sports}, Tables~\ref{tab:emb_health} and Tables~\ref{tab:emb_science}, Across all domains (\textsc{Sports}, \textsc{Health}, \textsc{Science}), 
we observed consistent trends aligned with our main findings: 
diagonal values in the GRADE matrix exhibit a steady increase from upper-left to bottom-right. 
This consistency across encoders demonstrates that the framework’s evaluation signals 
are not artifacts of a particular embedding choice, 
but rather reflect intrinsic aspects of multi-hop reasoning difficulty.

\subsection{Evaluation under Alternative Chunking Strategies}

To examine whether the GRADE framework is sensitive to specific retrieval granularities, 
we evaluated two alternative chunking strategies in addition to the main configuration: 
\texttt{64--128 tokens, 25-token overlap} and 
\texttt{256--512 tokens, 100-token overlap}. 

As shown in Table~\ref{tab:token_sports}, Table~\ref{tab:token_health} and Table~\ref{tab:token_science}, across all domains (\textsc{Sports}, \textsc{Health}, \textsc{Science}), 
we observed the same characteristic patterns as in our main experiments: 
diagonal values in the GRADE matrix exhibit a steady increase from upper-left to bottom-right. 
This indicates that the validity of the framework’s difficulty signals does not depend on 
a particular chunking configuration. 

\begin{table}[t]
\centering
\small
\begin{tabular}{lcccc}
\toprule
\textbf{Hops} & \textbf{Low $D_r$} & Q2 & Q3 & \textbf{High $D_r$} \\
\midrule
\multicolumn{5}{c}{\texttt{256--512 tokens, 100-token overlap}} \\
2-hop & \textbf{0.278} & 0.315 & 0.204 & 0.407 \\
3-hop & 0.140 & \textbf{0.388} & 0.673 & 0.340 \\
4-hop & 0.190 & 0.286 & \textbf{0.381} & 0.372 \\
5-hop & 0.314 & 0.529 & 0.543 & \textbf{0.429} \\
\midrule
\multicolumn{5}{c}{\texttt{64--128 tokens, 25-token overlap}} \\
2-hop & \textbf{0.222} & 0.259 & 0.204 & 0.333 \\
3-hop & 0.120 & \textbf{0.490} & 0.673 & 0.400 \\
4-hop & 0.286 & 0.333 & \textbf{0.619} & 0.488 \\
5-hop & 0.429 & 0.529 & 0.714 & \textbf{0.486} \\
\bottomrule
\end{tabular}
\caption{Error rates in the \textsc{Sports} domain with alternative chunking strategies.}
\label{tab:token_sports}
\end{table}

\begin{table}[t]
\centering
\small
\begin{tabular}{lcccc}
\toprule
\textbf{Hops} & \textbf{Low $D_r$} & Q2 & Q3 & \textbf{High $D_r$} \\
\midrule
\multicolumn{5}{c}{\texttt{256--512 tokens, 100-token overlap}} \\
2-hop & \textbf{0.373} & 0.224 & 0.433 & 0.672 \\
3-hop & 0.310 & \textbf{0.351} & 0.466 & 0.603 \\
4-hop & 0.327 & 0.429 & \textbf{0.490} & 0.580 \\
5-hop & 0.455 & 0.455 & 0.432 & \textbf{0.682} \\
\midrule
\multicolumn{5}{c}{\texttt{64--128 tokens, 25-token overlap}} \\
2-hop & \textbf{0.403} & 0.239 & 0.537 & 0.642 \\
3-hop & 0.310 & \textbf{0.386} & 0.517 & 0.586 \\
4-hop & 0.367 & 0.531 & \textbf{0.531} & 0.660 \\
5-hop & 0.455 & 0.523 & 0.500 & \textbf{0.727} \\
\bottomrule
\end{tabular}
\caption{Error rates in the \textsc{Health} domain with alternative chunking strategies.}
\label{tab:token_health}
\end{table}

\begin{table}[t]
\centering
\small
\begin{tabular}{lcccc}
\toprule
\textbf{Hops} & \textbf{Low $D_r$} & Q2 & Q3 & \textbf{High $D_r$} \\
\midrule
\multicolumn{5}{c}{\texttt{256--512 tokens, 100-token overlap}} \\
2-hop & \textbf{0.246} & 0.333 & 0.475 & 0.689 \\
3-hop & 0.350 & \textbf{0.417} & 0.483 & 0.672 \\
4-hop & 0.228 & 0.491 & \textbf{0.456} & 0.526 \\
5-hop & 0.352 & 0.415 & 0.519 & \textbf{0.574} \\
\midrule
\multicolumn{5}{c}{\texttt{64--128 tokens, 25-token overlap}} \\
2-hop & \textbf{0.213} & 0.350 & 0.475 & 0.672 \\
3-hop & 0.383 & \textbf{0.400} & 0.567 & 0.525 \\
4-hop & 0.246 & 0.456 & \textbf{0.368} & 0.544 \\
5-hop & 0.315 & 0.547 & 0.537 & \textbf{0.574} \\
\bottomrule
\end{tabular}
\caption{Error rates in the \textsc{Science} domain with alternative chunking strategies.}
\label{tab:token_science}
\end{table}

\clearpage

\subsection{Prompts}
\label{appendix:prompts}

We include all prompt templates used throughout the pipeline in this appendix. These prompts cover claim generation, entity matching, triple extraction, multi-hop question construction, and system evaluation.

\newpage
\label{app:prompt-design}
\subsubsection{Prompt for Claim Generation}

\begin{minipage}[h]
{\linewidth}\raggedright
\setlength{\parindent}{0cm}
\hrule
\vspace{1mm}

\small{
\texttt{\textcolor{blue}{System:}} A **claim** is a statement or assertion made within a text that expresses a belief, opinion, or fact. Given the evidence and the original context, please transform the evidence into a claim.

\hspace{0.5cm}

Note:

- The claim should be a clear and concise statement that logically follows from the provided evidence.

- The claim should not contain ambiguous references such as "he," "she," or "it." Use complete names or specify entities where necessary.

- The claim must be a paraphrased version of the evidence, stating the point or fact clearly, without adding extra information.

- If there is no claim that can be drawn from the evidence, please leave the response blank.

\hspace{0.5cm}

\texttt{\textcolor{blue}{User:}} 

Context: \{context\}

Evidence: \{evidence\}

Claim:  \newline
\textcolor{gray}{\# Content}
}
\vspace{1mm}
\hrule
\end{minipage}

\vfill

\subsubsection{Prompt for Claim-Sentence Consistency Check}

\begin{minipage}[h]
{\linewidth}\raggedright
\setlength{\parindent}{0cm}
\hrule
\vspace{1mm}

\small{
\texttt{\textcolor{blue}{System:}} You are an AI assistant that receives pairs of sentences and claims. \\
Your task is to determine whether each claim is consistent with its corresponding sentence. \\
Focus solely on whether the claim accurately reflects the core factual content of the sentence. \\
Ignore style, tone, attitude, or figurative language. \\
Respond with "Yes" if the claim is factually consistent with the sentence. \\
Respond with "No" if the claim introduces information that is not supported or is inconsistent. \\
Output format: Yes / No \\

\hspace{0.5cm}

\texttt{\textcolor{blue}{User:}} 

Sentence: {sentence}

Claim: {claim} \newline
\textcolor{gray}{\# Content}
}
\vspace{1mm}
\hrule

\end{minipage}

\vfill

\newpage

\subsubsection{Prompt for Triple Extraction}
 
\begin{minipage}[h]
{\linewidth}\raggedright
\setlength{\parindent}{0cm}
\hrule
\vspace{1mm}

\small{
\texttt{\textcolor{blue}{System:}} You are an AI assistant that extracts entities and their relationships from a list of sentences.
Each sentence has an associated sentence ID. \\

Your task is to extract triplets from each sentence in the form of:
(source\_entity|relationship|target\_entity|sentence\_id) \\

\hspace{0.5cm}

Please follow these guidelines: \\
- An entity can be a person, place, object, concept, or any meaningful noun phrase that participates in a relationship. \\
- Extract all valid (source\_entity|relationship|target\_entity) triplets from each sentence. \\
- Append the sentence ID at the end of each triplet to indicate which sentence it came from. \\
- If multiple triplets can be extracted from a single sentence, list all of them. \\
- Do not include duplicate triplets where only the order of source and target is reversed. \\

\hspace{0.5cm}

IMPORTANT: Resolve pronouns \\
- Replace pronouns such as he, she, it, they, this, that with the most specific entity mentioned in the sentence. \\

Output format: \\
(source\_entity|relationship|target\_entity|1)  \\
(source\_entity|relationship|target\_entity|2)  \\
(source\_entity|relationship|target\_entity|2)  \\
(source\_entity|relationship|target\_entity|3)  \\

\hspace{0.5cm}

\texttt{\textcolor{blue}{User:}}  \newline
Sentence 1: \{sentence1\} \\
Sentence 2: \{sentence2\} \\
... \\
Sentence 10: \{sentence10\} \\
\textcolor{gray}{\# Content}
}
\vspace{1mm}
\hrule
\end{minipage}

\vfill

\subsubsection{Prompt for Search Same Entity (Exact / Contextual Equivalence)}

\begin{minipage}[h]
{\linewidth}\raggedright
\setlength{\parindent}{0cm}
\hrule
\vspace{1mm}

\small{
\texttt{\textcolor{blue}{System:}} You are an AI assistant tasked with identifying entities that refer to the same concept based on a given set of triples and their supporting claims. \\

\hspace{0.5cm}

Each input consists of multiple (source\_entity, relationship, target\_entity) triples along with their corresponding claim context. \\
Your task is to group entities that can be considered the same, based on both the triples and their claim contexts. \\

\hspace{0.5cm}

There are two types of equivalence: \\
1. Always equivalent: Entities that refer to the same real-world object or concept in any context (e.g., "USA" and "United States"). \\
2. Context-dependent equivalent: Entities that refer to the same thing only in the context of the given triples and claim(s) (e.g., "study co-author" and "microplastics researcher"). \\

\hspace{0.5cm}

Format your output as follows: \\
Group identical entities together inside square brackets []. \\
Separate each entity with a vertical bar |. \\
At the end of each group, append either "always" or "context" (in quotes) to indicate the type of equivalence. \\
Write one group per line. \\
If no identical entities are found, output exactly: No identical entities found. \\

\hspace{0.5cm}

\texttt{\textcolor{blue}{User:}} \\
Example output: \\
\{examples\} \\

\hspace{0.5cm}

Triple: \{triples\} \\
Claim: \{claim\} \\
... \\
Triple: \{triples\} \\
Claim: \{claim\} \newline
\textcolor{gray}{\# Content}
}
\vspace{1mm}
\hrule
\end{minipage}

\newpage

\subsubsection{Prompt for QA Generation (Ground Truth)}

\begin{minipage}[h]
{\linewidth}\raggedright
\setlength{\parindent}{0cm}
\hrule
\vspace{1mm}

\small{
\texttt{\textcolor{blue}{System:}} You are an AI assistant designed to generate multi-hop questions and answers based on triples in the form of (source\_entity, relationship, target\_entity), along with the associated claims and context. \\

\hspace{0.5cm}

Your task is to generate a multi-hop question-answer pair based on the given triples. The number of hops should correspond to the number of triples provided. If there are N triples, generate a question that connects all N triples, and use them to form a coherent, logical path for the answer. \\

\hspace{0.5cm}

Ensure that: \\

- The question should begin with "Question:" and the answer should begin with "Answer:". \\
- The question should clearly reference the entities and relationships, and should be designed such that the answer is a concise, **specific entity or short phrase** (e.g., "Microsoft", "United States", "2025", "GLP-1 drugs"). \\
- The answer should **not be abstract** (e.g., "noticeable effects", "study participants", "potential limitations") but should be a **clear entity, specific term, or concise concept** that can be derived directly from the triples. \\
- The question and answer should be linked with a pipe (|) on the same line. \\
- Do not add external knowledge or assumptions beyond the given triples. \\

\hspace{0.5cm}

Notes for clarification: \\

\hspace{0.5cm}

- For N triples: The question should logically connect all N triples and form a coherent path that leads to a **specific, concrete answer** derived solely from the entities in the triples. \\
- Make sure the question is specific and each relationship in the chain is clearly traceable to lead to the final answer. \\

\hspace{0.5cm}

\texttt{\textcolor{blue}{User:}} \\
Example output format: \\
\{examples\} \\

\hspace{0.5cm}

Triples: \{triples\} \\
Claims: \{claims\} \\
Context: \{chunks\} \newline
\textcolor{gray}{\# Content}
}
\vspace{1mm}
\hrule
\end{minipage}

\newpage

\subsubsection{Prompt for Validation}

\begin{minipage}[h]
{\linewidth}\raggedright
\setlength{\parindent}{0cm}
\hrule
\vspace{1mm}

\small{
\texttt{\textcolor{blue}{System:}} You are an AI assistant tasked with reviewing question and answer pairs for ambiguity or vagueness. \\
Your goal is to evaluate whether each pair is clear and self-contained — that is, whether it can be understood without relying on external or missing context. \\

\hspace{0.5cm}

Use the following criteria to make your judgment:\  
The question and answer must be decontextualized — meaning they should be understandable on their own, without requiring additional background information. \\
If the answer includes vague references such as "other countries," "certain individuals," or "this technology," and the question does not provide enough information to specify what these refer to, then it is considered ambiguous. \\
Similarly, if the question uses pronouns or context-dependent expressions like "he," "they," "this," or "that" without clearly indicating the referent, the pair is not decontextualized and should be marked as ambiguous. \\

\hspace{0.5cm}

Based on these criteria: \\
If the question-answer pair is decontextualized and unambiguous, output True. \\
If it relies on missing context or includes vague or ambiguous expressions, output False. \\

\hspace{0.5cm}

Output format: \\
True / False \\

\hspace{0.5cm}

\texttt{\textcolor{blue}{User:}} \\
Question: \{question\} \\
Answer: \{answer\} \newline
\textcolor{gray}{\# Content}
}
\vspace{1mm}
\hrule
\end{minipage}

\newpage

\subsubsection{Prompt for Test RAG System Answer Generation}

\begin{minipage}[h]
{\linewidth}\raggedright
\setlength{\parindent}{0cm}
\hrule
\vspace{1mm}

\small{
\texttt{\textcolor{blue}{System:}} You are an AI assistant designed to generate answers for multi-hop questions. Given a question and its corresponding context, use only the information in the context to generate a **specific, concise answer**. \\

\hspace{0.5cm}

The answer should be **a clear, short entity, concept, or term**, such as "Microsoft", "United States", or "2020". Do not provide detailed explanations or longer sentences.  \\

\hspace{0.5cm}

Do not use any external knowledge or make assumptions. Focus solely on the information provided in the context to answer the question. \\

\hspace{0.5cm}

Output format: \\
Answer \\

\hspace{0.5cm}

\texttt{\textcolor{blue}{User:}} \\
Question: \{question\} \\
Context: \{top\_chunks\} \newline
\textcolor{gray}{\# Content}
}
\vspace{1mm}
\hrule
\end{minipage}

\subsubsection{Prompt for LLM Evaluation}

\begin{minipage}[h]
{\linewidth}\raggedright
\setlength{\parindent}{0cm}
\hrule
\vspace{1mm}

\small{
\texttt{\textcolor{blue}{System:}} You are an AI assistant that receives a question along with two answers: a ground truth answer and a generated response. Your task is to evaluate whether the generated response is correct or not, and provide a binary judgment (True or False). \\ 

\hspace{0.5cm}

Output format: \\
True/False

\hspace{0.5cm}

\texttt{\textcolor{blue}{User:}} \\
Question: \{question\} \\
Ground Truth Answer: {gt\_answer} \\
Response: \{rag\_answer\} \newline
\textcolor{gray}{\# Content}
}
\vspace{1mm}
\hrule
\end{minipage}

\end{document}